# XGBOD: Improving Supervised Outlier Detection with Unsupervised Representation Learning

Yue Zhao
University of Toronto
PricewaterhouseCoopers
Toronto, Canada
yuezhao@cs.toronto.edu

Maciej K. Hryniewicki
Data Assurance & Analytics
PricewaterhouseCoopers
Toronto, Canada
maciej.k.hryniewicki@pwc.com

*Abstract*—A new semi-supervised ensemble algorithm called XGBOD (Extreme Gradient Boosting Outlier Detection) is proposed, described and demonstrated for the enhanced detection of outliers from normal observations in various practical datasets. The proposed framework combines the strengths of both supervised and unsupervised machine learning methods by creating a hybrid approach that exploits each of their individual performance capabilities in outlier detection. XGBOD uses multiple unsupervised outlier mining algorithms to extract useful representations from the underlying data that augment the predictive capabilities of an embedded supervised classifier on an improved feature space. The novel approach is shown to provide superior performance in comparison to competing individual detectors, the full ensemble and two existing representation learning based algorithms across seven outlier datasets.

## I. INTRODUCTION

Outlier detection methods are widely used to identify anomalous observations in data [1]. However, using supervised outlier detection is not trivial, as outliers in data typically constitute only small proportions of their encompassing datasets. In addition, unlike traditional classification methods, the ground truth is often unavailable in outlier detection [2]–[4]. For supervised algorithms, such highly imbalanced datasets and insufficiently labeled data have led to limited generalization capabilities of these methods [2]. Over the years, numerous unsupervised algorithms have been developed for outlier detection. These methods specialize in exploring outlier-related information such as local densities, global correlation and hierarchical relationships for unlabeled data.

Ensemble methods combine multiple base classifiers to create algorithms that are more robust than their individual counterparts [5]. In the past several decades, numerous ensemble frameworks have been proposed, such as bagging [6], boosting [7] and stacking [8]. Although ensemble methods have been explored for both supervised and unsupervised applications, outlier ensemble techniques have been rarely studied [3]. As outlier detection algorithms are typically unsupervised and lack true labels, their construction is not trivial [2], [4]. Most existing outlier ensemble methods are unsupervised, using either bagging approaches such as Feature Bagging [9] or boosting approaches such as SELECT [3]. However, the predictive capabilities of supervised methods are often far too reliant on the proportion of labelled data that may exist within the dataset. Therefore, stacking-based outlier ensembles may be used to leverage both the label-related information using supervised learning as well as the complex data representations with unsupervised outlier methods.

The research presented herein extends and improves the work of Micenková et al. [10], [11] and Aggarwal et al. [12] to propose a semi-supervised ensemble framework for outlier detection. The original feature space is augmented by applying various unsupervised outlier detection functions on itself. Transformed outlier scores (TOS) generated by unsupervised outlier detection functions are viewed as richer representations of the data. Greedy TOS selection algorithms are then applied to prune the augmented feature space, in order to control the computational complexity and improve the accuracy of the prediction. Finally, the supervised ensemble method XGBoost [13] is used as the final output classier on the refined feature space. This combination of original features with the outputs of various base unsupervised outlier detection algorithms allows for a better representation of the data, similar to the classification meta-framework stacking [8].

The motivation behind this research is that unsupervised outlier detection algorithms are better at learning complex patterns in extremely imbalanced datasets than supervised methods. The strategy of taking the output of unsupervised methods as the input to the supervised classifier is regarded as a process of representation learning [10], [11] or unsupervised feature engineering [12]. Stacking is used as a combination framework to learn the weights of original features and newly generated TOS automatically. Compared to existing works [10], [11], the approach presented in this research does not rely on the costly EasyEnsemble [14] method to handle data imbalance by building multiple balanced samples; rather, it uses XGBoost [13] instead. In addition, multiple TOS selection methods are designed, evaluated and compared to achieve an efficient computational budget. Moreover, XGBOD does not require data pre-processing in feature combination, i.e., feature scaling for logistic regression in [10], [11], leading to an easier setup. Lastly, a tentative theoretical explanation of XGBOD is provided under a recently proposed framework by Aggarwal and Sathe [15]. Overall, XGBOD is easy to use, efficient to implement, and empirically effective for outlier detection.

## II. RELATED WORK

### A. Representation Learning

The effectiveness of a machine learning algorithm relies heavily on selected data representations or features [16], wherein abundant and effective representations tend to produce good prediction results. Some machine learning algorithms, such as deep learning, are capable of learning both the mapping from representations to outputs as well as the representations themselves. However, these algorithms require a large amount of data to extract useful representations, which is not typically available in outlier mining. Nonetheless, the concept is easily transferable to outlier detection: unsupervised outlier detection methods could be viewed as instruments to extract richer representations from limited data, which is also known as unsupervised feature engineering [12]. This approach has been proven to be effective in enriching the data expression and improving the supervised learning [17].

### B. Data Imbalance and Extreme Gradient Boosting

Data imbalance occurs when a given class (or subset) within the dataset represents an underwhelming minority of the overall population of classes [18]. When data imbalance is present, the performance of a classifier is generally degraded. Outlier detection is a binary classification task that is relatively imbalanced [15]. The outliers are inherently the minority class, which makes the detection of outliers prohibitively difficult. To treat data imbalance, bootstrap aggregating (Bagging) or EasyEnsemble methods are involved in outlier detection [10], [11]. EasyEnsemble builds multiple balanced subsamples by down-sampling the majority class, and combines the base classifier outputs trained on subsamples, such as majority vote. However, these methods are expensive to execute and their performance is problem-specific [19].

Extreme Gradient Boosting, commonly referred to as XGBoost, is a tree-based ensemble method developed by Chen [13]. It is a scalable and accurate implementation of gradient boosted trees, explicitly designed for optimizing the computational speed and model performance. Compared to established boosting algorithms like gradient boosting, XGBoost utilizes a regularization term to reduce the over-fitting effect, yielding better predictions [13] and shorter execution times [20]. Recent research shows that ensemble methods with XGBoost have the greatest ability to handle imbalanced datasets relative to other ensemble methods [18]. As a result, XGBoost is selected as the final supervised classifier to replace EasyEnsemble in this study. Additionally, XGBoost could automatically generate feature importance rankings while fitting the data [20], which is useful for implementing a feature pruning scheme to improve the computational efficiency of the algorithm presented herein.

### C. Unsupervised Outlier Detection Methods

Unsupervised methods do not rely on label information and could learn outlier characteristics through various approaches, such as local density. Developed unsupervised outlier detection methods may be categorized into four groups [21]: (i) linear models such as Principal Component Analysis; (ii) Proximity-Based Outlier models, including density- or distance-based methods; (iii) statistical and probabilistic models that rely on value analysis; and (iv) high-dimensional outlier models such as Isolation Forest. These models are based on different assumptions, and yield superior results on certain datasets when the corresponding assumptions are met. In this research, various types of unsupervised outlier detection methods are used as base detectors to construct an effective ensemble.

### D. Outlier Ensemble

Numerous ensemble methods have been introduced previously in the context of outlier detection [2], [4], [12]. These studies have either combined the outputs of constituent detectors or induced the diversity among different constituent detectors with potentially independent errors [3]. The most straightforward combination strategy is averaging the outputs of various base detectors after the normalization, also known as Full Ensemble. One of the earliest works, Feature Bagging [9], induces diversity by building on a randomly selected subset of features. Rayana and Akoglu adapt the boosting approach into outlier mining [3]. Their algorithm, SELECT, generates pseudo label information to perform sequential learning [3]. It is noteworthy that these frameworks are unsupervised, and similar to bagging and boosting methods in traditional classification tasks. In this study, we combine the results of various unsupervised detectors via a similar supervised approach called Stacking [8]. Stacking has been used recently in combing supervised and unsupervised ensembles in knowledge base population tasks [22]. As mentioned by Aggarwal et al. in [12], the key difference between classification stacking and the outlier Stacking presented in this research is that the former utilizes supervised methods for representation learning, whereas this approach relies on unsupervised methods instead. In both cases, the final output classier however is supervised.

### E. Semi-supervised Outlier Ensemble with Feature Learning

Micenková et al. have proposed a semi-supervised framework called BORE to leverage the strength of both supervised and unsupervised methods [10], [11]. BORE first uses various unsupervised outlier detection methods to generate outlier scores on the training data. These unsupervised outlier scores are then combined with the original features to construct the new feature space. To combat the data imbalance in outlier data, they use EasyEnsemble [14] to create multiple balanced training samples and they then average the results of the samples. Logistic regression with $L2$ regularization is applied on the subsamples to identify outliers. To measure the computational cost, they simulate a cost-aware feature selection that takes the randomly generated running cost into account. Similar to Micenková's work, Aggarwal and Sathe discuss a semi-supervised outlier ensemble framework that uses mild supervision in combining with unsupervised outlier models [12]. With access to a small number of outlier labels, they state that one could use logistic regression or support vector machines to learn the weights of base detectors. $L1$ regularization is suggested to prevent overfitting and perform feature selection while many detectors are presented. We include both BORE (logistic regression with $L2$ regularization) and logistic regression with $L1$ regularization in this study as baseline algorithms.

## III. ALGORITHM DESIGN

XGBOD is a three-phase framework, as depicted in Fig. 1. In the first phase, new data representations are generated. Specifically, various unsupervised outlier detection methods are applied to the original data to get transformed outlier scores as new data representations. In the second phase, a selection process is performed on newly generated outlier scores to keep the useful ones. The selected outlier scores are then combined with the original features to become the new feature space. Finally, an XGBoost classifier is trained on the new feature space, and its output is regarded as the prediction result.

### A. Phase I: Unsupervised Representation Learning

The proposed approach is based on the notion that unsupervised outlier scores can be viewed as a form of learned representations of the original data [10]–[12]. Alternatively, these can also be understood as a form of unsupervised feature engineering, to augment the original feature space as well.

Let the original feature space $X \in \mathbb{R}^{n \times d}$ denote a set of $n$ data points with $d$ features. As outlier detection is a binary classification, vector $y \in \{0,1\}$ assigns outlier labels, where 1 represents outliers and 0 represents normal points. Let $L$ be a set of labeled observations of $X$, such that:

$$L = \{(x_1, y_1), ..., (x_n, y_n)\} \in \mathbb{R}^{n \times d} \quad (1)$$

The outlier scoring function is defined as a mapping function $\Phi(\cdot)$, where each scoring function would output a real-valued vector $\Phi_i(X) \in \mathbb{R}^{n \times 1}$ on dataset $X$ as the **transformed outlier scores (TOS)** to describe the degree of outlyingness. Outlier scoring functions could be any unsupervised outlier detection method. The outputs, TOS, are used as new features to augment the original feature space. Combining $k$ outlier scoring functions together constructs a transformation function matrix: $\Phi = [\Phi_1, ..., \Phi_k]$ which generates the outlier score matrix of $k$ base scoring functions on the original feature space $X$. Applying $\Phi(\cdot)$ on the original data $X$, the outlier scoring matrix $\Phi(X)$ is then given as:

$$\Phi(X) = \left[\Phi_1(X)^T, ..., \Phi_k(X)^T\right] \in \mathbb{R}^{n \times k} \quad (2)$$

As mentioned above, any unsupervised outlier detection method could be used as a base outlier scoring function for feature transformations. However, heterogeneous base functions tend to yield better results, as identical outputs from base functions do not greatly contribute to the ensemble [1], [12]. The diversity among base functions encourages distinct data characteristics to be learned, leading to an improved generalization ability of the ensemble. Furthermore, highly correlated base estimators result in similar errors and do not contribute to the prediction; rather, they bring unnecessarily high computational burdens to the overall solution.

Meanwhile, outlier scoring functions should be accurate as well, as inaccurate ones degrade the prediction. As a result, there is an inherent tradeoff between diversity and accuracy:

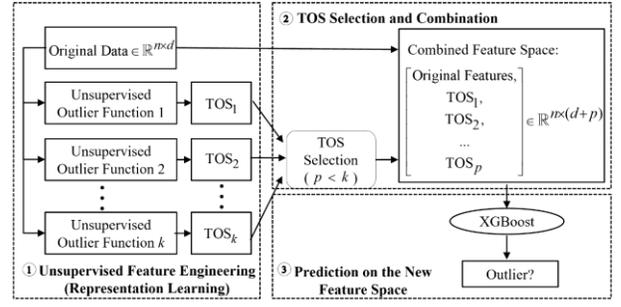

Fig. 1. Illustration of XGBOD's semi-supervised approach

using distinct but inaccurate detectors improves the diversity at the risk of degrading the overall predictive capabilities. Therefore, a balance between diversity and accuracy should be maintained to gain an improved prediction result [1], [4]. In this study, different types of unsupervised outlier methods are used as the base outlier scoring functions, and their parameters are also tweaked to generate further variation. This design yields a varied collection of both accurate and inaccurate TOS.

### B. Phase II: TOS Selection

Once the outlier score matrix $\Phi(X)$ is generated, it is ready to be combined with the original features $X$. In the work of Micenková et al. [11], $X$ is directly combined with the entire newly generated data representations $\Phi(X)$ as:

$$\text{Feature Space}_{new} = [X, \Phi(X)] \in \mathbb{R}^{n \times l} \quad (3)$$

where $l = (d + k)$ is the dimension of the combined feature space with $d$ original features and $k$ newly generated TOS.

Compared to the approach presented herein, multiple TOS selection methods are designed to pick only $p$ ($p \leq k$) TOS from $\Phi(X)$ for combining with the original features. The reason is closely related to feature selection in machine learning: not all TOS would contribute to the prediction. Additionally, reducing the number of selected TOS would speed up the execution as (i) there are fewer transformations to apply on the original data and (ii) the combined feature spaces are smaller to learn on. Three selection methods are defined, and an empty set $S$ is initialized to store the selected TOS.

**Random Selection** picks $p$ TOS from $\Phi(X)$ randomly and adds to $S$ without replacement.

**Accurate Selection** selects the top $p$ most accurate TOS. The accuracy measure could be any appropriate evaluation metric, such as the area under receiver operating characteristic curve (ROC), among others. Let $ACC_i(\cdot)$ denote the ROC of $\Phi_i(X)^T$ measured by the ground truth $y$. It then iteratively selects the most accurate TOS in $\Phi(X)$ based on the value of $ACC_i(\cdot)$ using:

$$ACC_i = ROC(\Phi_i(X)^T, y) \quad (4)$$

**Balance Selection** maintains the balance between diversity and accuracy by picking the TOS that are both accurate and diverse. For each $\Phi_i(X) \in \Phi(X)$, a greedy selection is executed based on TOS accuracy calculated by Eq. (4). To improve diversity in $S$ at the same time, a discounted accuracy function $\Psi(\Phi_i)$ is devloped on the basis of Eq. (4) as:

$$\Psi(\Phi_i) = \frac{ACC_i}{\sum_{j=1}^{\#(S)} |\rho(\Phi_i, \Phi_j)|} \quad (5)$$

subject to $\Phi_j \in \{S\}, ACC_i \geq 0$

The Pearson correlation $\rho(\Phi_i, \Phi_j)$ is used to measure the correlation between a pair of TOS. The Pearson correlation between a TOS and all selected TOS in $S$ is aggregated as $\sum_{j=1}^{\#(S)} |\rho(\Phi_i, \Phi_j)|$. If a TOS is highly correlated to the TOS that have already been selected in $S$, a larger denominator would be assigned in Eq. (5) to discount its accuracy. The discounted accuracy function favors accurate TOS that have low correlations with already selected TOS in $S$, thereby discouraging the with-in set similarity in $S$. Until the size of set $S$ equals $p$, TOS in $\Phi(X)$ will be evaluated iteratively by Eq. (5). Each time, the TOS with the largest discounted accuracy $max(\Psi(X)_i)$ is added to $S$ and removed from the candidate pool $\Phi(X)$. The workflow is given in Algorithm 1.

With running one of the TOS selection algorithms above, $p$ TOS are selected as $S \in \mathbb{R}^{n \times p}$. Then the refined feature space, Feature Space$_{comb}$, is created by concatenating the original features $X$ by $S$, as Feature Space$_{comb} = [X, S] \in \mathbb{R}^{n \times (d+p)}$. It is noted that $(k - p)$ TOS are discarded to improve the algorithm efficiency and prediction.

*C. Phase III: Prediction with XGBoost*

An XGBoost classifier is applied on Feature Space$_{comb}$ to generate the final output. Leveraging XGBoost, the run-time efficiency and predictive capability of the algorithm are enhanced due to its robustness to data imbalance and overfitting. Additionally, a post-pruning process may be performed by XGBoost's internal feature importance. The feature importance is calculated on the feature counts in node splitting, when the model is fitted. More aggressive TOS pruning is thus possible, i.e. selecting top $q$ most important TOS from $S$ by the internal feature ranking.

*D. Theoretical Foundations*

Bias-Variance tradeoff is widely used to understand the generalization error of a classification algorithm. Recently, Aggarwal and Sathe have pointed out that a similar theoretical framework is also applicable to outlier ensemble [15]. In this view, an outlier ensemble has two types of reducible errors: (i) squared bias, caused by limited ability to fit the data and (ii) variance, caused by the sensitivity to the training data. An effective outlier ensemble should successfully control the reducible error, given that reducing bias may increase variance and vice versa.

In this research, various unsupervised outlier detection algorithms are used to enrich the feature space, which injects diversity into the model and then combines the results. This is considered as a variance reduction approach, as combining diverse base detectors reduces the variance of outlier ensemble [3], [12], [15]. However, this may cause inaccurate TOS to be included in the ensemble, incurring higher model bias. This explains why the full ensemble (averaging all TOS) does not perform well—it may include some inaccurate base detectors with high bias [3]. Thus, the TOS design selection algorithms only keep the useful ones for reducing the bias. Moreover, the ensemble and regularization mechanisms in XGBoost could achieve low variance without introducing much bias [20]. With various instruments to reduce bias and variance, XGBOD is considered to improve the generalization ability in all stages. However, the performance of XGBOD may be heuristic and unpredictable with pathological datasets or a bad selection of base unsupervised outlier detection functions.

## IV. EXPERIMENT DESIGN

Two comparison analyses are conducted with various baselines included. ROC [1], [10], [11] and Precision@n (P@N) [12] are widely used for evaluation in outlier detection; both are included herein. The final scores are calculated by averaging the results of 30 independent trials. To further compare the performance difference, various statistical tests are introduced. Specifically, the non-parametric Friedman test [23] followed by Nemenyi post-hoc test [24] is used to analyze the differences among multiple algorithms, while Wilcoxon ranksum test is used to conduct the pairwise comparisons. In this study, $p < 0.05$ is regarded as statistically significant.

---

**Algorithm 1** Balance Selection

**Input**: $\Phi = \{\Phi_1, ..., \Phi_k\}$, ground truth $y$, # of TOS $= p$
**Output**: The Set of Selected TOS: $S$
**Initialize**: Selected TOS: $S = \{\}$

1. $\Phi(X)_{max} = max(ACC(\Phi(X)))$ /* most accurate*/
2. $S \leftarrow S \cup \Phi(X)_{max}$ /*add selected TOS to set $S$ */
3. $\Phi(X) \leftarrow \Phi(X) \setminus \Phi(X)_{max}$ /*remove from the pool*/
4. **while** $\#(S) < p$ **do**
5.    **for** $\Phi(X)_i \in \Phi(X)$ **do**
6.       $\Psi(\Phi_i) \leftarrow$ Eq. (5) /*discounted accuracy*/
7.       $\Phi(X)_{max} = max(\Psi(X)_i)$
8.       $S \leftarrow S \cup \Phi(X)_{max}$ /*add the current best to set $S$ */
9.       $\Phi(X) \leftarrow \Phi(X) / \Phi(X)_{max}$ /*remove from the pool*/
10.    **end for**
11. **end while**
12. **return** $S$

TABLE I. SUMMARY OF DATASETS

| Dataset | Points ($n$) | Features ($d$) | Outliers |
|---|---|---|---|
| Arrhythmia | 452 | 274 | 66 (15%) |
| Letter | 1600 | 32 | 100 (6.25%) |
| Cardio | 1831 | 21 | 176 (9.6%) |
| Speech | 3686 | 600 | 61 (1.65%) |
| Satellite | 6435 | 36 | 2036 (31.64%) |
| Mnist | 7603 | 100 | 700 (9.21%) |
| Mammography | 11863 | 6 | 260 (2.32%) |

### A. Outlier Datasets

Table I summarizes seven real-world datasets used in this study. All datasets are widely used in outlier research [2], [11], [15], [25], and are publicly accessible within an outlier detection repository called *Odds* [26]. The datasets are split into training sets (60%) and testing sets (40%).

### B. Base Outlier Scoring Function and Parameter Setting

The effectiveness of XGBOD depends on both the accuracy and diversity of the base outlier scoring functions. Therefore, a wide range of unsupervised outlier detection algorithms are included. The pruning process then selects the most useful ones. In this research, seven base outlier scoring functions are used: (i) *k*NN (the Euclidean distance of the *k*th nearest neighbor as the outlierness score); (ii) *k*-Median; (iii) Avg-*k*NN (average *k* nearest neighbor distance as the outlierness score); (iv) LOF [27]; (v) LoOP [28]; (vi) One-Class SVM [29] and (vii) Isolation Forests [25]. The description of these algorithms is omitted due to brevity. However, it is noted that *k*NN, *k*-Median, Avg-*k*NN and One-Class SVM herein are all unsupervised with no requirements of the ground truth. To further induce diversity, the parameters of these base scoring functions are varied. For nearest neighbor-based algorithms including *k*NN, *k*-Median, Avg-*k*NN, LOF and LoOP, the range of *k* is defined as $[1, 2, 3, 4, 5, 10, 15,..., 100]$. Given that the LoOP algorithm is computationally expensive on large datasets, the narrower *k* range of $[1, 3, 5, 10]$ is used. For One-Class SVM, the kernel is fixed to radial basis function, and different upper bounds on the fraction of training errors are used. For Isolation Forest, the number of base estimators varies at $[10, 30, 50, 70, 100, 150, 200, 250]$. In total, 107 TOS is built for each dataset.

### C. Experiment Setting

**Experiment I** compares the performance of different frameworks. They use either all TOS or none TOS (on the original data); no TOS selection is included. For example, in the proposed method, *XGB_Comb*, applies XGBoost directly on the combined data constructed by the original features and all TOS. Multiple baselines are included: (i) *Best_TOS*: the highest score among all TOS (unsupervised); (ii) *Full_TOS*: the average of all TOS scores, equivalent to Full Ensemble (unsupervised); (iii) *XGB_Orig*: XGBoost on the original data with no TOS; (iv) *XGB_New*: XGBoost on newly generated TOS only; (v) *L2_Comb*: *L*2 logistic regression with EnsyEnsemble on the combined data, also known as BORE in [10] and (vi) *L1_Comb* [12]: *L*1 logistic regression with EnsyEnsemble on the combined data. Both *L2_Comb* and *L1_Comb* are using EasyEnsemble (50 bags). XGBoost in both Experiment I and II uses 100 base estimators with max tree depth at 3 by default.

**Experiment II** analyzes the effect of TOS selection. XGBoost is used as the only classifier in all settings. Therefore, selecting zero TOS is equivalent to *XGB_Orig* and selecting all TOS is equivalent to *XGB_Comb* in Experiment I. From selecting zero TOS (*XGB_Orig*) to all TOS (*XGB_Comb*), it would result in distinct feature spaces. The results of selecting different numbers of TOS are analyzed. Besides the number of selected TOS, the results of different selection algorithms (Random Selection, Accurate Selection and Balance Selection) are also compared. It is noted that selecting one TOS with Accurate Selection is same as using Balance Selection.

## V. RESULTS AND DISCUSSIONS

### A. Prediction Performance Analysis

Table II shows the results of Experiment I, which directly compares different classification methods without any TOS selection. For clarity, all unsupervised approaches are marked with *, and all methods applied on the combined feature space are marked with # in Table II. The best performer for each dataset is highlighted in bold.

The Friedman test illustrates that there is a statistically significant difference among seven algorithms for both ROC ($\chi^2 = 32.45$, $p < 0.001$) and P@N ($\chi^2 = 32.88$, $p < 0.001$). However, Nemenyi test is not strong enough to show pairwise differences. *XGB_Comb* achieves the best result on six out of seven datasets except **Mammography**, for both ROC and P@N. For **Mammography**, *XGB_Comb* is inferior to *XGB_Orig*; incorporating TOS does not improve the result in this case. One potential explanation is that **Mammography** only has 6 features so that not all chosen unsupervised methods can extract useful representations out of such a limited feature space. In contrast, unsupervised representation learning markedly improves the supervised classifier on datasets with a high dimension feature space, such as **Arrhythmia** (274 features), **Mnist** (100 features) and **Speech** (600 features). Therefore, unsupervised representation learning tends to be more useful when a high feature dimension is presented.

It is noted that the final output classifier should be carefully chosen while using unsupervised representation learning. One view is that simple linear models, such as logistic regression, should be used as the final classifier for the augmented feature space. In this situation, EasyEnsemble should be used to construct balanced subsamples, since simple model could not handle the data imbalance. In contrast, we argue that the combination of a simple model and EasyEnsemble could be replaced by the ensemble algorithms with strong regularization, like XGBoost, for better performance. The experiment result confirms our view. *XGB_Comb*, *L1_Comb* and *L2_Comb* are all applied on the combined feature space (the original features + newly generated TOS). However, only *XGB_Comb* has an edge over the supervised approach with no TOS (*XGB_Orig*), while *L1_Comb* and *L2_Comb* are even inferior to *XGB_Orig* on **Arrhythmia**, **Cardio, Satellite** and **Mnist**.

TABLE II. MODEL PERFORMANCE (AVERAGE OF 30 TRIALS)

| Datasets | ROC | | | | | | | P@N | | | | | | |
|---|---|---|---|---|---|---|---|---|---|---|---|---|---|---|
| | Best_TOS* | Full_TOS* | L1_Comb# | L2_Comb# | XGB_Orig | XGB_New | XGB_Comb# | Best_TOS* | Full_TOS* | L1_Comb# | L2_Comb# | XGB_Orig | XGB_New | XGB_Comb# |
| Arrhythmia | 0.8288 | 0.7750 | 0.8537 | 0.8545 | 0.8698 | 0.8110 | **0.8816** | 0.5449 | 0.4088 | 0.5530 | 0.5449 | 0.5932 | 0.4568 | **0.6002** |
| Letter | 0.9368 | 0.8897 | 0.9653 | 0.9685 | 0.9399 | 0.9593 | **0.9729** | 0.5655 | 0.3450 | 0.6653 | 0.6874 | 0.6181 | 0.6679 | **0.7320** |
| Cardio | 0.9449 | 0.8411 | 0.9953 | 0.9879 | 0.9966 | 0.9868 | **0.9976** | 0.5972 | 0.3478 | 0.8925 | 0.8508 | 0.9302 | 0.8477 | **0.9377** |
| Speech | 0.7673 | 0.5009 | 0.8515 | 0.8534 | 0.7593 | 0.7819 | **0.8591** | 0.1677 | 0.0441 | 0.1569 | 0.2082 | 0.1696 | 0.1502 | **0.2561** |
| Satellite | 0.7534 | 0.6992 | 0.9156 | 0.9096 | 0.9656 | 0.9254 | **0.9666** | 0.6188 | 0.5040 | 0.7566 | 0.7508 | 0.8508 | 0.7769 | **0.8568** |
| Mnist | 0.9184 | 0.8526 | 0.9890 | 0.9880 | 0.9963 | 0.9980 | **0.9999** | 0.4368 | 0.4147 | 0.8409 | 0.8379 | 0.9195 | 0.9302 | **0.9901** |
| Mammography | 0.8759 | 0.8617 | 0.9410 | 0.9415 | **0.9515** | 0.9105 | 0.9431 | 0.3152 | 0.2609 | 0.6013 | 0.5754 | **0.6877** | 0.4707 | 0.6677 |

TABLE III. TRAINING AND PREDICTION EFFICIENCY (SECONDS)

| Dataset | XGB_Orig | XGB_New | XGB_Comb | L1_Comb | L2_Comb |
|---|---|---|---|---|---|
| Arrhythmia | 0.3963 | 0.2693 | **0.6181** | 1.5521 | 0.6281 |
| Letter | 0.1520 | 0.9048 | 1.0286 | 1.2231 | **0.5719** |
| Cardio | 0.1456 | 0.1046 | 1.1174 | 3.1053 | **0.6215** |
| Speech | 8.1853 | 2.2833 | 10.177 | 5.9067 | **2.2505** |
| Satellite | 0.5018 | 3.7720 | **4.1570** | 144.79 | 28.252 |
| Mnist | 1.6107 | 3.5495 | **4.9213** | 19.359 | 26.428 |
| Mammography | 0.2968 | 5.2578 | 5.4165 | 38.182 | **1.9436** |

Compared with *L1_Comb* [12] and *L2_Comb* [10], *XGB_Comb* outperforms regarding both ROC and P@N; it brings more than 10% improved P@N on four datasets and even 23.05% improvement on **Speech**. It has been found that *L1_Comb* and *L2_Comb* have close performance, confirmed by the Wilcoxon rank-sum test on *L1_Comb* and *L2_Comb* with no statistically significant difference.

The analysis of execution time is conducted on selected algorithms, and the TOS generation stage is not measured since it depends strongly on the choice of base algorithm and the implementation. Table III shows the execution time (average of 30 trials), and the most efficient method with the full combined feature space is highlighted in bold. Clearly, *L1_Comb* is generally least efficient and markedly slower than *L2_Comb* and *XGB_Comb*; it is not recommended to use for this task. However, there is no clear winner between *L2_Comb* and *XGB_Comb* regarding running efficiency. *L2_Comb* is faster on four datasets and 2 to 3 times faster than *XGB_Comb*, while *XGB_Comb* is the most efficient algorithm on three datasets that is 5 to 7 times faster than *L2_Comb*. It is noted *XGB_Comb* is the most efficient one on time consuming datasets (**Satellite** and **Mnist**). Additionally, the authors have not measured the data preprocessing time for *L1_Comb* and *L2_Comb*, such as data scaling, which is not required for *XGB_Comb*. Combining with the analysis of the detection performance, it is assumed that *XGB_Comb* is still a stable choice that brings consistent efficiency improvement over *L1_Comb* and meaningful improvement over *L2_Comb* on time-consuming tasks.

*B. Number of Selected TOS*

Figure 2 illustrates the XGBoost's P@N performance when different numbers of TOS are used to enrich the original feature space, from zero (*XGB_Orig*) to all TOS (*XGB_Comb*). Detailed performance metrics are omitted for brevity. In general, using a subset of TOS usually generates better results than using all TOS. For instance, the best P@N is achieved by selecting 5 TOS on **Arrhythmia**, 60 TOS on **Letter**, 5 TOS on **Cardio**, 80 on **Satellite** and 5 on **Mammography**. Using all TOS on **Mammography** even leads to degraded performance which is worse than no TOS (*XGB_Orig*). In contrast, if 5 most accurate TOS (*Accurate_5*) are selected, the prediction performance is improved. Surprisingly, combining only a small number of TOS with the original feature may still improve the results significantly. For instance, using the most accurate TOS (*Accurate_1*) would improve P@N from 0.6181 to 0.7307 on **Letter**, and improve from 0.7673 to 0.8338 on **Speech**. This observation can be explained by the design of XGBoost: it learns the important features automatically by identifying most frequent features to split on in base trees [20]. Although only a small number of useful TOS included, XGBoost assigns higher weights to these key features in prediction.

Using t-distributed stochastic neighbor embedding (t-SNE) [30], two-dimensional visualizations of **Arrhythmia** are presented with four distinct feature spaces in Fig. 3. Augmenting the original feature space with 10 TOS (upper right) and 30 TOS (lower left) improves the data expression over the original feature space (upper left), since outliers (red triangles) are more separated from normal points (blue dots) with TOS. The visualization of using TOS only (lower right) demonstrates that TOS are good representations of the data where outliers are easier to be identified. However, deciding the best number of TOS to use is non-trivial and possibly data-dependent; Friedman Test does not show a statistical difference regarding the number of TOS. It is understood that selecting too few TOS is risky with high model variance, which can be controlled by selecting more TOS. As a rule of thumb, including all TOS is a safe choice; it often results in decent, if not the best, performance.

Besides using the combined feature space (the original feature space along with the newly generated TOS), using newly generated TOS alone achieves excellent results occasionally. On **Letter** and **Speech** data, using TOS alone outperforms the original feature space. This does not only prove that TOS can effectively express the data but also implies that the original features may not be necessary for the final classifier. However, this performance improvement is inconsistent. For example, although Fig. 3 suggests using TOS alone results in better data representation, using TOS alone has lower ROC than using the original feature space on **Arrhythmia**. The same phenomenon has been observed in [10], [11] as well. Thus, the original features cannot be fully replaced by TOS as the quality of TOS is heuristic in the current approach. Despite, the possibility still exists if effective representation extraction methods can be proposed.

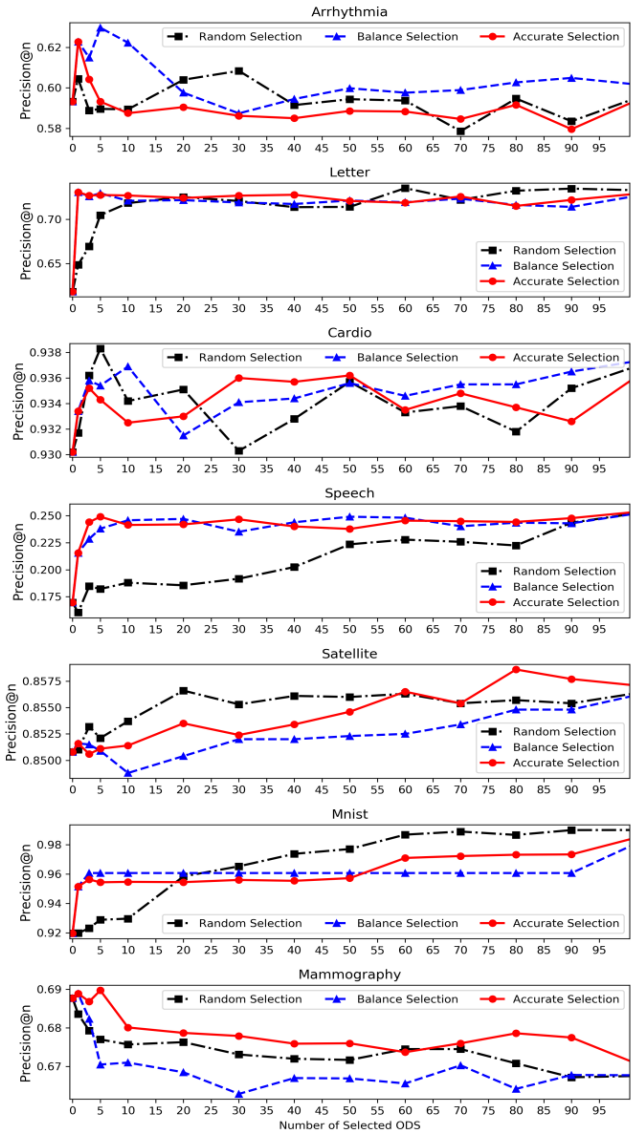

Fig. 2. The effect of number of TOS and selection method

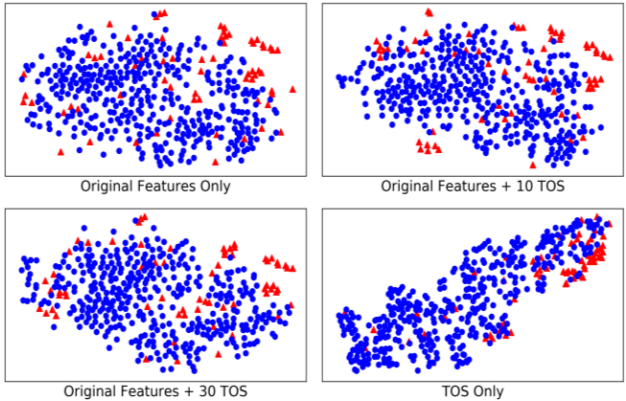

Fig. 3. t-SNE visualization on **Arrhythmia**

## C. TOS Selection Method

Figure 2 also shows the comparison of three selection methods regarding P@N. On **Letter**, **Cardio** and **Mnist**, the Friedman test does not show a significant difference. On the remaining four datasets, the selection method performances tend to be more arbitrary with significant differences, although no algorithm consistently outperformed.

Random Selection comes with more uncertainty. Its performance is severely degraded on **Speech** by selecting only one TOS (0.1604), even worse than not using TOS (0.1696). Additionally, Random Selection performs worse than Accurate Selection and Balance Selection on all datasets if only selecting one TOS. However, Random Selection also has promising results; it ranks highest on **Cardio** (10 TOS) and **Mnist** (90 TOS). Micenková's work [10] analyzed the performance of Random Selection and found it is consistently poor. The results presented herein agree that Random Selection is less predictable and stable, but it could occasionally lead to decent results. The authors assume the discrepancy is because (i) EasyEnsemble in [10], [11] is less stable than built-in bagging of XGBoost and (ii) different outlier datasets have distinct characteristics. While more research is needed, it is possible that Random Selection exhibits a heuristic result.

Choosing from Balance Selection and Accurate Selection may depend on the dimension of the original feature space. Empirically, Balance Selection seems to work better on the datasets with more features, such as **Speech** (600 features) and **Arrhythmia** (274 features). In contrast, Accurate Selection outperforms on **Mammography** (6 features) and **Satellite** (36 features). One assumption is Accurate Selection tends to pick TOS generated by a specific type of outlier detection method with different parameters, but it is hard for this type of method to extract distinct but useful representations on high dimensional data. Alternatively, Balance Selection favors diversity, leading to good results by selecting different types of TOS with independent errors. Accurate Selection might be appropriate when the original feature space dimension is low, as emphasis on diversity might not be critical in this situation.

With the number of selected TOS increases, all three selections methods become more comparable because the number of overlapping TOS rises as well. As discussed, Random Selection is riskier, and its performance is usually the worst of all three algorithms. It is not recommended especially while selecting few TOS. However, it may give superior results sometimes due to its heuristic nature. Accurate Selection is suggested when the dimension of the original features is low, and Balance Selection might be useful with a complex original feature space. Using all TOS like the default setting of *XGB_Comb* could be a safe choice in most of the cases.

## D. Limitations and Future Directions

Numerous studies are underway. Firstly, TOS are extracted from the original features directly. However, making feature selection on the original data may eliminate some unnecessary ones, and TOS can then be built on the selected features. Secondly, more TOS selection methods can be incorporated in future studies, such as post-pruning on XGBoost's feature importance. Additionally, TOS Selection may be replaced by dimensionality reduction methods, like principal components analysis; TOS could be combined instead of being selected.

## VI. Conclusions

A new semi-supervised outlier ensemble method, XGBOD (Extreme Gradient Boosting Outlier Detection), has been proposed, described and demonstrated for the detection of outliers in various benchmarked datasets. XGBOD is a three-phase system that (i) uses unsupervised outlier detection algorithms to improve data representation (ii) leverages greedy selection to keep useful representations and then (iii) applies an XGBoost classifier to predict on the improved feature space. Numerical experiments on seven outlier datasets show that XGBOD achieves markedly improved results compared to competing approaches. This is supported by the theoretical considerations that indicate reductions in variance and bias.

The design of XGBOD is motivated by the previous work of Micenková et al. [10], [11] and Aggarwal et al. [12], which proposed that unsupervised outlier detection methods can extract richer outlier data representations than the original feature space. Specifically, applying various established unsupervised outlier detection algorithms on the original data could generate TOS with potentially better representations. Furthermore, combining these TOS with the original feature space could then improve the overall outlier prediction.

This research extends these previous studies to show that using even very few TOS significantly improves the outlier detection rate. The t-SNE visualizations on different feature spaces confirm that TOS help to separate outliers from normal observations. To control the computational expense, three TOS selection algorithms have been designed and tested herein. Recommendations regarding the selection, use and interpretation of these algorithms have also been provided. In general, Balance Selection is proposed for high-dimensional feature spaces and Accurate Selection is suggested for data with fewer features. Random Selection may be useful in some cases; however, the result is generally unpredictable.

Compared to other semi-supervised outlier ensemble methods, XGBOD provides better predictive capabilities, eliminates the dependency of building balanced subsamples and averaging the results, and improves efficiency with more stable execution. It is robust enough to handle an increased range of input features, in that it does not require any feature scaling or missing value imputation in data preprocessing. To the authors' best knowledge, XGBOD is the first complete framework that combines unsupervised outlier representation with supervised machine learning methods that use ensemble trees. Lastly, it should be noted that all source codes, datasets and figures used in this study are openly shared and available[1].

---

[1] https://github.com/yzhao062/XGBOD